\documentclass[tinyml,hidelinks]{acmart}
\usepackage{times}

\usepackage{soul}
\usepackage{url}
\usepackage{hyperref}
\usepackage[utf8]{inputenc}
\usepackage[font=small,skip=5pt]{caption}
\usepackage{graphicx}

\usepackage{booktabs}
\usepackage[ruled,vlined]{algorithm2e}
\usepackage{verbatim}
\usepackage{algorithmic}
\usepackage[T1]{fontenc}    
\usepackage{mathtools}      
\usepackage{amsfonts}       
\usepackage{nicefrac}       
\usepackage{microtype}      
\usepackage{subfigure}
\usepackage{multirow}
\usepackage{tikz}
\usepackage{pgfplots}
\pgfplotsset{compat=1.15}
\usepackage{float}
\usepackage{rotating}
\usepackage{color}
\usepackage{makecell}
\usepackage{boxedminipage}
\usepackage{flushend}
\DeclareMathOperator*{\argmin}{argmin}

\AtBeginDocument{%
  \providecommand\BibTeX{{%
    \normalfont B\kern-0.5em{\scshape i\kern-0.25em b}\kern-0.8em\TeX}}}

\setcopyright{rightsretained}
\copyrightyear{2023}
\acmYear{2023}

\begin{document}

\title{Automatic Network Adaptation for Ultra-Low Uniform-Precision Quantization}

\author{Seongmin Park}
\email{skstjdals@hanyang.ac.kr}
\authornote{Both authors contributed equally to this research.}
\affiliation{%
  \institution{Hanyang University}
  \city{Seoul}
  \country{Republic of Korea}
}

\author{Beomseok Kwon}
\email{qjatjr9913@gmail.com}
\authornotemark[1]
\authornote{This work was done while the author was at Hanyang University}
\affiliation{%
  \institution{NAVER CLOVA}
  \city{Seongnam}
  \country{Republic of Korea}}

\author{Kyuyoung Sim}
\email{kysim@nota.ai}
\affiliation{%
  \institution{Nota AI}
  \city{Seoul}
  \country{Republic of Korea}}

\author{Jieun Lim}
\authornote{This work was done while the author was at Nota AI}
\email{jieun.lim@sapeon.com}
\affiliation{%
  \institution{SAPEON Korea Inc.}
  \city{Seongnam}
  \country{Republic of Korea}}

\author{Tae-Ho Kim}
\email{thkim@nota.ai}
\affiliation{%
  \institution{Nota AI}
  \city{Seoul}
  \country{Republic of Korea}}

\author{Jungwook Choi}
\email{choij@hanyang.ac.kr}
\authornote{Corresponding author}
\affiliation{%
  \institution{Hanyang University}
  \city{Seoul}
  \country{Republic of Korea}
}

\renewcommand{\shortauthors}{Park and Kwon, et al.}

\begin{abstract}
Uniform-precision neural network quantization has gained popularity since it simplifies densely packed arithmetic unit for high computing capability.
However, it ignores heterogeneous sensitivity to the impact of quantization errors across the layers, resulting in sub-optimal inference accuracy. This work proposes a novel neural architecture search called \textit{neural channel expansion} that adjusts the network structure to alleviate accuracy degradation from ultra-low uniform-precision quantization. The proposed method selectively expands channels for the quantization sensitive layers while satisfying hardware constraints (e.g., FLOPs, PARAMs). Based on in-depth analysis and experiments, we demonstrate that the proposed method can adapt several popular networks' channels to achieve superior 2-bit quantization accuracy on CIFAR10 and ImageNet. In particular, we achieve the best-to-date Top-1/Top-5 accuracy for 2-bit ResNet50 with smaller FLOPs and the parameter size.
\end{abstract}


\maketitle

\section{Introduction}
Deep neural networks (DNNs) have reached human-level performance in a wide range of domains including image processing (\cite{tan2019efficientnet}), object detection (\cite{tan2020efficientdet}), machine translation (\cite{devlin2018bert}), and speech recognition (\cite{nassif2019speech}). However, the tremendous computation and memory costs of these state-of-the-art DNNs make them challenging to deploy on resource-constrained devices such as mobile phones, edge sensors, and drones. Therefore, several edge hardware accelerators specifically optimized for intensive DNN computation have emerged, including Google's edge TPU(\cite{googletpu}) and NVIDIA's NVDLA (\cite{nvdla}).  

One of the central techniques innovating these edge DNN accelerators is the quantization of deep neural networks (QDNN). QDNN reduces the complexity of DNN computation by quantizing network weights and activations to low-bit precision. Since the area and energy consumption of the multiply-accumulate (MAC) unit can be significantly reduced with the bit-width reduction (\cite{sze2017efficient}), thousands of them can be packed in a small area. Therefore, the popular edge DNN accelerators are equipped with densely integrated low-precision MAC arrays to boost their performance in compute-intensive operations such as matrix multiplication (MatMul) and convolution (Conv).

Early studies of QDNN focus on the quantization of weights and activations of MatMul and Conv to the same bit-width (\cite{hubara2016binarized,zhou2016dorefa}). This \textit{uniform-precision quantization} gained popularity because it simplifies the dense MAC array design for DNN accelerators. However, uniform bit allocation does not account for the properties of individual layers in a network. \cite{sakr2018analytical} shows that the optimal bit-precision varies within a neural network from layer to layer. As a result, uniform-precision quantization may lead to sub-optimal inference accuracy for a given network.

Mixed-precision quantization addresses this limitation by optimizing bit-widths for each layer. In this approach, the sensitivity of the layer to the quantization error is either numerically estimated (\cite{dong2019hawq}) or automatically explored under the framework of neural architecture search (NAS, \cite{cai2020rethinking}) to allocate bit-precision properly.
However, mixed-precision networks require variable precision support in hardware, restricting computation units' density and power efficiency (\cite{camus2019review}). Therefore, mixed-precision support imposes a significant barrier for the low-profile edge accelerators with stringent resource budgets.

In this work, we propose \textbf{a novel NAS based automatic network adaptation method that can address the layer-wise heterogeneous sensitivity under ultra-low uniform-precision quantization}. The proposed method explores network structure in terms of the number of channels. Different from the previous work that only includes pruning of the channels in its search space (\cite{dong2019network}), we further incorporate the expansion of the channels, thus called \textit{neural channel expansion} (NCE). During a search of NCE, search parameters associated with different numbers of channels are updated based on each layer's sensitivity to the uniform-precision quantization and the hardware constraints such as FLOPs and the number of parameters (PARAMs). The more sensitive to quantization errors, the larger number of channels preferred in that layer. When the preference to the larger number of channels in a layer exceeds a certain threshold, we expand the channels in that layer's search space to explore the more number of channels. Therefore, NCE allows both pruning and expansion of each layer's channels, finding the sweet-spot for the trade-off between the robustness against the quantization error and the hardware cost.  

Based on in-depth analysis and experiments, we demonstrate that NCE can facilitate the search to adapt the target model's structure for better accuracy under ultra-low uniform precision. The experimental results of challenging 2-bit uniform quantization on CIFAR10 and ImageNet show that the network structures adapted from the popular convolutional neural networks (CNNs) achieve superior accuracy while maintaining the same level of (or lower) hardware costs. In particular, we achieve the best-to-date accuracy of 74.03/91.63\% (Top-1/Top-5) for NCE-ResNet50 on ImageNet with slightly lower FLOPs and 30\% reduced PARAMs.

Our contributions can be summarized as follows:
\begin{itemize}
\itemsep0em
\item We propose a new NAS-based quantization algorithm called \textit{neural channel expansion} (NCE), which is equipped with a simple yet innovative channel expansion mechanism to balance the number of channels across the layers under uniform-precision quantization.
\item We provide in-depth analysis to show that NCE's search space is capable of finding a network structure more robust to quantization errors than the previous channel adaptation techniques such as WRPN(\cite{mishra2018wrpn}) and TAS(\cite{dong2019network}).
\item We demonstrate that the proposed method can adapt the structure of target neural networks to significantly improve the quantization accuracy while maintaining the same level of (or lower) hardware costs. 
\end{itemize}

\section{Related Work}

\textbf{Neural architecture search}: 
The goal of NAS is to find a network architecture that can achieve the best test accuracy. Early studies (\cite{zoph2016neural}) often employed meta-learners such as reinforcement learning (RL) agents to learn the policy for accurate network architectures. However, RL-based approaches may incur prohibitive computation costs (e.g., thousands of GPU hours). As a relaxation, a differentiable neural architecture search (DNAS) has been proposed (\cite{liu2018darts}), which updates the architecture parameters and the weights via bi-level optimization. Recent DNAS approaches considered hardware constraints such as FLOPs and PARAMs to explore the trade-off between the network's accuracy and hardware efficiency. For example, TAS (\cite{dong2019network}) proposed a DNAS-based channel pruning method to discover a resource-efficient neural network architecture. This work extends the DNAS framework toward an automatic adjustment of neural network channels with improved robustness against uniform-precision quantization errors.

\textbf{Low-precision quantization of deep neural network}: 
Early work on QDNN (\cite{hubara2016binarized,zhou2016dorefa}) introduced the concept of a straight-through estimator (STE) for the approximation of gradients of the non-differentiable rounding operation. This approximation enabled uniform-precision (1- or multi-bit) quantization during model training to fine-tune the weight parameters for robustness against the quantization error. QDNN techniques have evolved to adaptively find the quantization step size (PACT~\cite{choi2018pact}, LSQ~\cite{esser2020learned}), which significantly enhanced the accuracy of the uniform-precision quantization. However, this line of research lacks consideration of the heterogeneous quantization sensitivity for individual layers in a network. On the other hand, mixed-precision quantization allows layer-specific bit-precision optimization; the higher bit-precision is assigned to the more quantization sensitive layers. \cite{dong2019hawq} numerically estimated the sensitivity via approximating the impact of quantization errors on model prediction accuracy. \cite{wang2019haq} employed a reinforcement learning framework to learn the bit-allocation policy. EdMIPS (\cite{cai2020rethinking}) adopted DNAS with the various bit-precision operators in the search space. 
However, mixed-precision representation requires variable precision arithmetic units with complex data path and controls (\cite{camus2019review}). Therefore, mixed-precision support incurs significant overhead in area and power consumption of the DNN acceleration hardware. 
In this work, we propose a novel NAS based network adaptation to capture the layer-wise heterogeneous quantization sensitivity while utilizing uniform-precision operations for efficient hardware support.

\textbf{Neural-Net adaptation for accurate DNN quantization}:
\label{subsec:channel-expansion}
Researchers have actively studied the adaptation of neural network architecture for accurate DNN quantization. WRPN (\cite{mishra2018wrpn}) demonstrated that increasing channels of the neural network helped regain QDNN accuracy, but with a quadratic increase of FLOPs and PARAMs. \cite{zhao2019improving} further attempted to selectively split the channels with large magnitude weights in the pre-trained models. This channel splitting reduced the dynamic range of weights to be represented with lower bit-precision (6- to 8-bits). 
However, it is not straightforward to extend this numerical remedy of dynamic range to ultra-low bit QDNN.
In this work, we propose a novel neural-net adaptation method that can selectively expand channels via DNAS to decrease the dynamic range when uniform-precision quantization is applied. The proposed method can improve the accuracy of ultra-low (e.g., 2-bit) uniform-precision quantization while maintaining original FLOPs and PARAMs via selective channel adjustment. 

\section{Neural Channel Expansion}
\label{sec:method}
In this section, we explain the detail of our neural channel expansion (NCE) method. In NCE, the goal is to find a network structure defined by the architecture parameters associated with each layer $\alpha=\{\alpha_{l}\}_{l=1:L}$ that minimizes the validation loss $\mathcal{L}_{val}$ of a QDNN with a uniform $b$ bit-precision $\mathcal{N}_{Q(b)}$ and the weights $\mathcal{W}$, as shown in Eq.~(\ref{eq:optimize}).

\begin{equation}
\begin{aligned}
\label{eq:optimize}
\alpha^* = \argmin_{\alpha}  \mathcal{L}_{val}(\mathcal{N}_{Q(b)}(\alpha,W_\alpha)) \qquad \\
\text{s.t.}\; W_\alpha = \argmin_W \mathcal{L}_{train}(\mathcal{N}_{Q(b)}(\alpha,W)), \qquad  \\   
\mathcal{L}_{val} = \mathcal{L}_{CE} +  \lambda_{flop}\mathcal{L}_{flop}+\lambda_{param}\mathcal{L}_{param}. 
\end{aligned}
\end{equation}

Note that the validation loss consists of the cross-entropy loss as well as the loss considering hardware constraints (FLOPs, PARAMs). We solve this optimization problem via bi-level optimization of DNAS (\cite{liu2018darts}). Similar to TAS (\cite{dong2019network}), we construct the search space over the number of channels $C=\left \{1:c_{out}  \right \}$ with the search parameters $\alpha_l \in \mathbb{R}^{|C|}$. Then the output activation $\hat{O}$ is computed as the weighted sum of sampled activations with a different number of channels aligned via channel-wise interpolation (CWI): 
\begin{equation}
\begin{aligned}
\hat{O}=\sum_{j\in I}^{}Softmax(\alpha_j;\{\alpha_k\}_{k\in I}) \\
\times CWI(O_{1:C_j}, max\{c^k_{out}\}_{k\in I}), 
\end{aligned}
\end{equation}
where $O_{j:1\leq j\leq c_{out}} =\sum_{k=1}^{c_{in}} Q(X\{k,:,:\})*Q(W\{j,k,:,:\})$ is computed with input activation $X$ and weight $W$ quantized by the quantizer $Q$, and $I$ is the sampled subset of $C$.  

In TAS, the number of channels ($|C|$) is fixed, limiting the exploration scope to the pruning. In NCE, we enable channel expansion of individual layers when the search parameter associated with the maximum number of channels exceeds the channel preference threshold, $T$. The intuition is that if one layer is susceptible to the quantization errors, its search parameters are updated toward the preference for a larger number of channels to decrease the cross-entropy loss. 
With this expansion condition, we can expand channels to those layers affected most by the quantization errors and prune channels of the other layers robust to quantization; therefore, the overall hardware constraints are met.

\begin{algorithm}[ht]
    \caption{{\bf Neural Channel Expansion} \label{Algorithm}}

\KwIn{}
\quad Split the training set into two dis-joint sets: $D_{weight}$ and $D_{arch}$ ($n(D_{weight}) = n(D_{arch})$) \\
\quad Search Parameter: \{$\alpha_1^l, \alpha_2^l, .., \alpha_n^l$\} $\in A^l$, \,\,  \{$A^1, A^2, ..,A^L$\} $\subset  \mathbb{A}$, \,\, $L=$number of layer \\
\quad Expand Threshold: $T$ \\   

\nl {\bf For} Warm-up Epoch {\bf do} \\
\nl \quad Sample batch data $D_w$ from $D_{weight}$ and network from $\mathbb{A} \sim U(0,1)$ \\
\nl \quad Calculate $Loss_{weight}$ on $D_w$ to update network weights \\
\nl {\bf End for}\\
\nl {\bf For} Search Epoch {\bf do} \\
\nl \quad Sample batch data $D_w$ from $D_{weight}$ and network from $Softmax(\mathbb{A})$\\
\nl \quad Calculate $Loss_{weight}$ on $D_w$ to update network weights \\
\nl \quad Sample batch data $D_a$ from $D_{arch}$ and network from $Softmax(\mathbb{A})$\\
\nl \quad Calculate $Loss_{arch}$ on $D_a$ to update $\mathbb{A}$ \\
\nl \quad {\bf For} layer {\bf do} \\
\nl \qquad $j\gets \# A^l$\\ 
\nl \quad \quad {\bf If} $Softmax(\alpha_j^l;\{\alpha_k^l\}_{k\in j}) \geq T$ {\bf do} \\

\nl \qquad \quad \textit{Expand} search space($\alpha_{j+1}^l$) \\
\nl \qquad \quad $\alpha_{j+1}^l \gets \alpha_j^l$  \quad {\bf\#} copy search parameter \\
\nl \qquad {\bf End if}\\
\nl \quad {\bf End for} \\
\nl {\bf End for} \\
\nl Derive the searched network from $\mathbb{A}$\\
\nl Randomly initialize the searched network and optimize it on the training set
\end{algorithm}

Algorithm~\ref{Algorithm} summarizes the overall procedure. NCE consists of three phases: warm-up, search, and train. 
As advocated by \cite{bender2020can}, we first perform a warm-up of the entire super-net so that all the super-net weight parameters can be reasonably initialized. The search phase consists of the iterative updates of weights ($W$) and the search parameters ($\alpha$) via bi-level optimization Eq.~(\ref{eq:optimize}). The updated search parameter associated with the maximum number of channels is compared with the threshold $T$ (pre-determined as a hyper-parameter) to identify the layers that need channel expansion. When channel expansion is performed (= \textit{Expand}), the additional weight parameters are added to that layer (and the search parameter is also copied), increasing the number of channels. Once the search is done, the candidate model is derived by the "winner-takes-all" strategy; i.e., for each layer, the number of channels with the largest magnitude search parameter is selected.

\section{Analysis}
\label{sec:analysis}
This section explains how NCE finds the network structures that are more robust to the uniform-precision quantization error while maintaining hardware constraints.

\subsection{Impact of channel expansion to quantization}
\label{subsec:dynamic-range}

\begin{figure}[t]
\begin{center}
\centerline{\includegraphics[width=\linewidth]{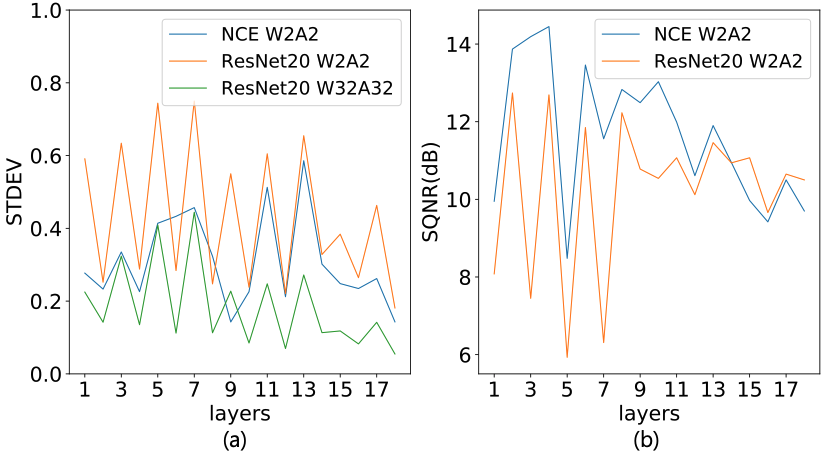}}
\caption{Comparison of (a) STDEV and (b) SQNR of activation between ResNet20-CIFAR10 and NCE when the models are quantized to 2-bit. (W2A2 Accuracy: ResNet20 (90.82\%), NCE (91.54\%)) }
\label{fig:analysis-STDEV}
\vspace{-0.5 cm}
\end{center}
\end{figure}

NCE expands channels to robustify the neural network against the quantization error. 
To understand the impact of channel expansion on DNN quantization, 
we first reveal that \textit{quantization applied to a given network substantially increases the dynamic range of activation, hindering successful DNN quantization}. Fig.\ref{fig:analysis-STDEV}a shows the standard-deviation (STDEV) of output activation of each layer for ResNet20 trained from scratch on CIFAR10, with and without quantization during training\footnote{This quantization-aware training follows the same hyper-parameter settings described in Sec.\ref{subsec:expr-setting}.}. W\{X\}A\{Y\} indicates that weights and activations are quantized into X- and Y-bits, respectively. There is a substantial increase of STDEV from W32A32 to W2A2, implying that the activation's dynamic range is increased due to 2-bit quantization. 
NCE expands the channels so that the increase in STDEV can be mitigated. As shown in Fig.\ref{fig:analysis-STDEV}b, this mitigation results in higher signal-to-noise-ratio (SQNR) than the quantization without NCE.  
Therefore, the network adapted by NCE can be more robust to quantization error.
 
\begin{figure}[t]
\begin{center}
\centerline{\includegraphics[width=\linewidth]{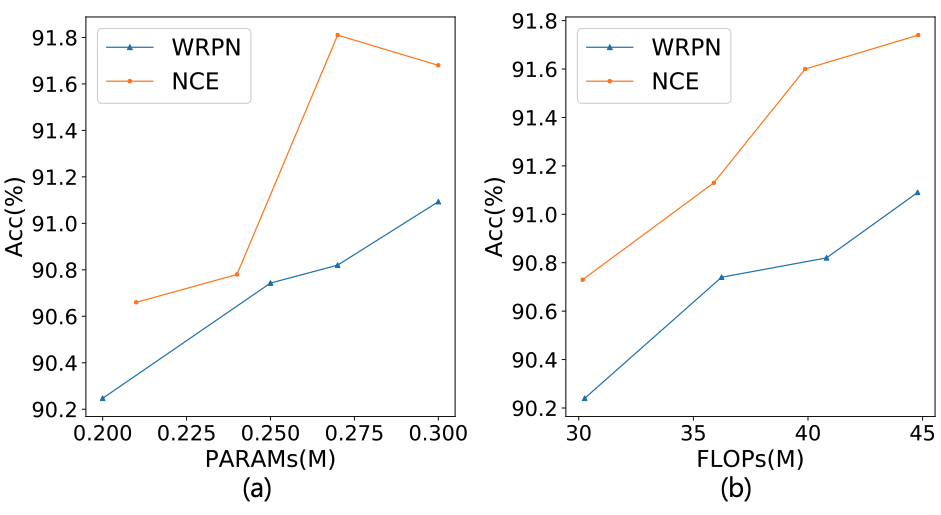}}
\caption{Comparison of uniform-quantization performance between NCE and WRPN with respect to (a) PARAMs and (b) FLOPs on 2-bit ResNet20-CIFAR10.}
\label{fig:nce-vs-wrpn}
\vspace{-0.5 cm}
\end{center}
\end{figure}

Channel expansion can increase the accuracy of QDNN, but at the cost of increased hardware cost (e.g., PARAMs and FLOPs). A straightforward channel expansion method is to uniformly increase the number of channels for all the layers, as discussed in WRPN (\cite{mishra2018wrpn}). We argue that NCE can explore a better trade-off between accuracy vs. hardware cost than WRPN by \textit{expanding the number of channels only for the necessary layers}. To validate this claim, we compare the trade-off curves between accuracy and the two popular hardware costs, PARAMs and FLOPs. As shown in Fig.~\ref{fig:nce-vs-wrpn}, NCE always achieves higher accuracy than WRPN for any PARAMs or FLOPs, confirming its superiority.

\subsection{Search space of neural channel expansion}
\label{subsubsec:benefit-channel-search}

\begin{figure}[t]
\begin{center}
\centerline{\includegraphics[width=0.8\linewidth]{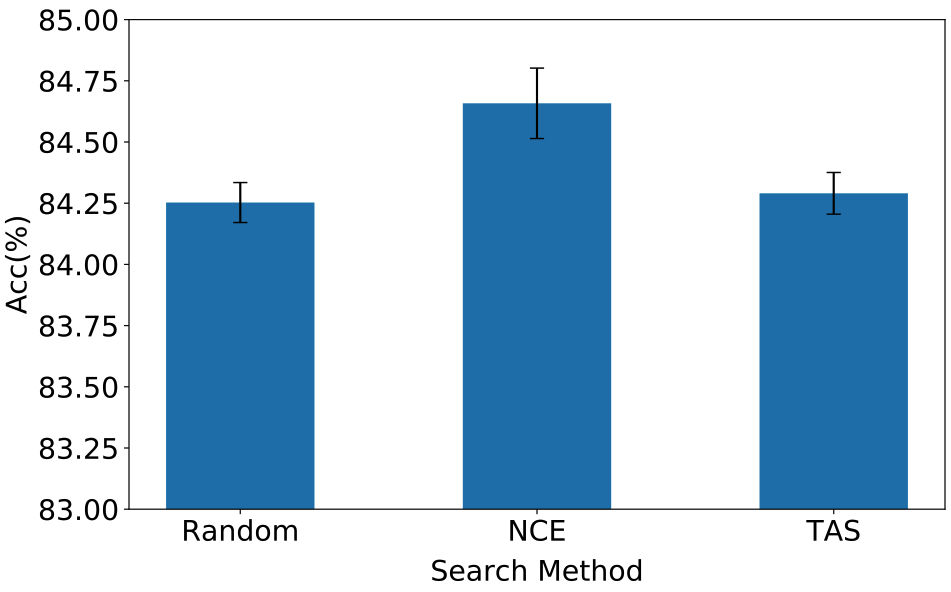}}
\caption{Comparison of channel search methods on 2-bit ResNet8-CIFAR10 model. Ten network structures are retrieved from each search method. Each network is trained from scratch for three times to measure test accuracy.}
\vspace{-0.5 cm}
\label{fig:cmpr-search-methods}
\end{center}
\end{figure}

The search space of NCE consists of the search parameters associated with the number of channels for each layer, where the search parameters of a layer can be expanded when a larger number of channels is preferred. \textit{This selective expansion of search space is the key to granting flexibility in architecture search.} To understand the benefit, we constructed a toy experiment with an 8-layer ResNet-CIFAR10, where the number of channels for seven Conv layers is explored under 2-bit uniform quantization. Three search methods are compared: 1) Random search (randomly select $\{0.75, 1, 1.25\}$ channel expansion ratio for each layer; total $3^7$ cases), 2) TAS search (prune the model with the maximum number of channels following \cite{dong2019network}), and 3) NCE. Fig.~\ref{fig:cmpr-search-methods} shows the averaged accuracy over ten searched neural network structures, each trained from scratched for three times. As shown in the figure, the 2-bit model accuracy found by TAS is marginally better than the models searched randomly, whereas NCE outperforms both Random search and TAS with a noticeable margin.
This result highlights the superiority of NCE's search space over TAS and Random search. 
In Sec.~\ref{subsec:analysis-ch-sel-pref}, we explain how NCE excels channel adaptation to better compensate the quantization errors.  


As NCE can expand search space for the layers that prefer a large number of channels, one may attribute the superiority of NCE to the enlarged search space. We argue that \textit{the benefit comes more from the flexibility of NCE's search space than from the search space size}. To confirm this claim, we constructed an experiment of 2-bit quantization on ResNet20-CIFAR10 with three search settings: 1) NCE starting from 1X channels accompanied by eight search parameters per layer, 2) TAS starting from 2X channels accompanied by eight search parameters, and 3) the same TAS with 16 search parameters. Note that NCE expands the search space up to 16 search parameters during the search, and thus NCE's search space is a subset of 2X TAS's search space. In Table~\ref{tab:selective-search}, we can observe that 2X TAS achieved only modest accuracy gain, whereas NCE achieved a noticeable accuracy gain with slightly lower FLOPs than 2X TAS. This experimental results demonstrate the benefit of NCE's selective channel expansion.  

\begin{table}[t]
\begin{center}
\resizebox{\linewidth}{!}{
\begin{tabular}{ccc}
\Xhline{2\arrayrulewidth} 
ResNet20 CIFAR10 W2A2               & Accuracy & FLOPs  \\ \hline 
Full-Prec Baseline                  & 92.88\%  & 40.81M \\ 
W2A2 Baseline                       & 90.82\%  & 40.81M \\ \hline
W2A2 2X TAS(\#search param=8)       & 90.99\%  & 43.11M \\ 
W2A2 2X TAS(\#search param=16)      & 91.12\%  & 40.36M \\ \hline
W2A2 NCE (\#search param=8$\sim$16) & 91.54\%  & 40.99M \\ \Xhline{2\arrayrulewidth} 
\end{tabular}}
\end{center}
\caption{Comparison of quantization performance between NCE and 2X TAS with controlled search space size.}
\vspace{-0.5 cm}
\label{tab:selective-search}
\end{table}

\subsection{Channel selection preference of NCE}
\label{subsec:analysis-ch-sel-pref}

\begin{figure}[t]
\begin{center}
\centerline{\includegraphics[width=\linewidth]{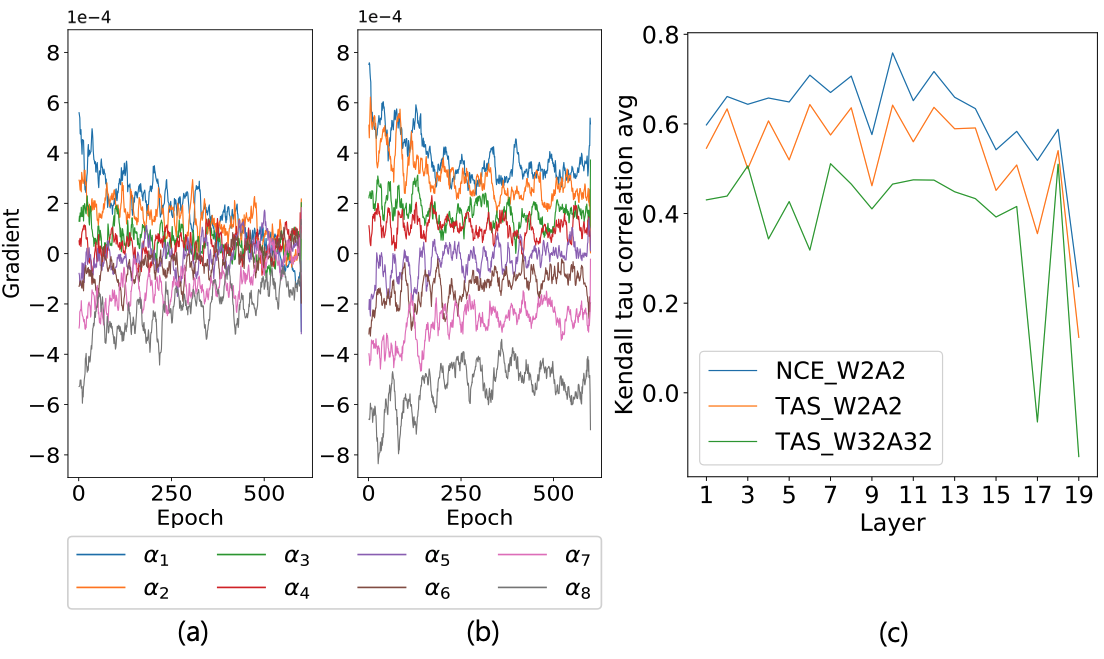}}
\caption{Experiments on ResNet20-CIFAR10 that shows gradients of cross-entropy loss w.r.t. search parameters ($\alpha_1 \sim \alpha_8$) of a layer during search: (a) in full-precision, (b) with 2-bit quantization. (c) Kendall rank-correlation score of all layers.}
\label{fig:analysis-grad}
\end{center}
\end{figure}

To find the reasoning behind the superiority of NCE's search space, we investigate channel selection preference during the search. 
Channel selection preference can be observed by the gradients w.r.t. the search parameters. 
As an example, Fig.~\ref{fig:analysis-grad}a shows the gradients of the search parameters during the TAS search of ResNet20-CIFAR10 in full-precision. Note that the search parameter associated with the maximum number of channels ($\alpha_8$) initially receives the negative gradients, whereas the search parameter with the least number of channels ($\alpha_1$) receives positive gradients\footnote{Since $\alpha = \alpha - \eta*grad_{\alpha}$, the negative gradients increase the magnitude of the search parameter.}. From this trend, we can conjecture that this layer initially prefers a large number of channels, but the preference diminishes over the epochs of search.

We conjecture that \textit{quantization during search excels the preference for a large number of channels}. As shown in Fig.~\ref{fig:analysis-grad}b, from the same experimental settings, if we apply quantization during the search, the preference to a large number of channels becomes more distinctive. This phenomenon is because the quantization errors affect the cross-entropy loss more. To quantify this channel preference, we calculate the \textit{Kendall rank correlation score} for the gradients of the search parameters and average over the search epochs; \textit{the more distinctive preference to a large number of channels, the higher the Kendall score}. 
Fig.~\ref{fig:analysis-grad}c shows the layer-wise Kendall score with and without quantization. Note that the Kendall score is increased if the quantization is applied during the architecture search. This increased Kendall score implies quantization drives the search parameters toward a strong preference to a larger number of channels.
In contrast to TAS's fixed search space, NCE allows expansion of search space, and thus, channel expansion can happen selectively for the layer with a strong preference for many channels. Interestingly, as shown in Fig.~\ref{fig:analysis-grad}c, NCE gains the higher Kendall score, indicating that \textit{NCE's search space expansion make preference to a large number of channels more distinctive}. 
Therefore, our simple yet novel search space of selective channel expansion makes channel adjustment more sensitive to the quantization errors, opening up opportunities for enhanced compensation of quantization errors.

\section{Experiments}
\label{sec:experiments}

\subsection{Experimental settings}
\label{subsec:expr-setting}
NCE is evaluated with popular CNNs trained on CIFAR10 and ImageNet datasets. We employ PACT (\cite{choi2018pact}) as the main quantization scheme. NCE is implemented in PyTorch based on the TAS framework\footnote{https://github.com/D-X-Y/AutoDL-Projects}. For CIFAR10 experiments, we conduct 200 epochs of warm-up followed by an NCE search for 600 epochs. We use the channel preference threshold ($T$) of 0.3 and the constraint coefficient of 2. For ImageNet experiments, we randomly choose 50 images from the original each 1000 classes to reduce the training time, similar to \cite{wu2019fbnet}. We conducted 40 epochs of warm-up followed by the search for 110 epochs. We used the channel preference threshold ($T$) of 0.19 and the constraint coefficient of 1.5. We optimized the weight via SGD and the architecture parameters via Adam. Regularization coefficient of PACT is 0.001. After the search, the candidate model is derived by the "winner-takes-all" strategy. Unless noted otherwise, all experiments on CIFAR10 are repeated three times, and the average test accuracy is reported. For a fair comparison with prior work (e.g., \cite{choi2018pact,jung2019learning}), the first/last and short-cut layers are not considered for 2-bit uniform-precision quantization. More detailed information about the experimental settings can be found in the supplementary materials.

\subsection{CIFAR10 results}

To evaluate NCE on CIFAR10, we employ three popular CNNs based on the ResNet structure and the other based on the VGG structure. For each network, we apply 2-bit uniform quantization with and without NCE. As shown in Table~\ref{tab:cifar10}, NCE consistently boosts the accuracy of QDNN w/o NCE by 0.35\% $\sim$ 0.72\% with the same level of (or lower) hardware costs (FLOPs and PARAMs). In particular, the network structure found by NCE has $11\sim 66\%$ lower parameters, implying that the accuracy improvement is because of the robustness of the adapted network structure rather than the model size. These experimental results demonstrate NCE's capability for automatic network adaptation for ultra-low uniform quantization.



\begin{table}[t]
\begin{center}
\resizebox{\linewidth}{!}{
\begin{tabular}{cclccc}
\Xhline{2\arrayrulewidth} 
Network                   & W32A32                   &         & W2A2             & FLOPs            & PARAMs         \\ \hline
\multirow{2}{*}{ResNet20} & \multirow{2}{*}{92.88\%} & w/o NCE & 90.82\%          & \textbf{40.81M}  & 0.27M          \\
                          &                          & w/ NCE  & \textbf{91.54\%} & 40.99M           & \textbf{0.24M} \\ \hline
\multirow{2}{*}{ResNet32} & \multirow{2}{*}{93.81\%} & w/o NCE & 92.22\%          & 69.12M           & 0.47M          \\
                          &                          & w/ NCE  & \textbf{92.64\%} & \textbf{66.63M}  & \textbf{0.42M} \\ \hline
\multirow{2}{*}{ResNet56} & \multirow{2}{*}{94.26\%} & w/o NCE & 93.08\%          & 125.75M          & 0.86M          \\
                          &                          & w/ NCE  & \textbf{93.43\%} & \textbf{123.04M} & \textbf{0.74M} \\ \hline
\multirow{2}{*}{VGG16}    & \multirow{2}{*}{94.24\%} & w/o NCE & 93.48\%          & 313.2M           & 14.72M         \\
                          &                          & w/ NCE  & \textbf{93.94\%} & \textbf{302.96M} & \textbf{5.01M} \\ \Xhline{2\arrayrulewidth} 
\end{tabular}}
\end{center}
\caption{CIFAR10: Comparison on test accuracy (average of 3-runs) and hardware costs for 2-bit QDNN with or without NCE.}
\label{tab:cifar10}
\end{table}

\subsection{ImageNet results}

We further evaluate NCE with ResNet structures on ImageNet by comparing it with state-of-the-art QDNN methods. We take reported Top-1 and Top-5 accuracy from the state-of-the-art uniform-precision quantization techniques; LSQ(\cite{esser2020learned}), QIL(\cite{jung2019learning}), 
and PACT(\cite{choi2018pact}). We also include the cutting edge mixed-precision quantization; EdMIPS(\cite{cai2020rethinking}) for comparison. The experimental results are summarized in Table~\ref{tab:imagenet}. Note that NCE consistently improves the accuracy of the original 2-bit quantized models (=w/o NCE) by large margins (1.67\% $\sim$ 2.09\%) thanks to the channel adaptation.
Furthermore, NCE mostly outperforms the accuracy of both uniform- and mixed-precision techniques for ResNet18 and ResNet50 with comparable (or lower) hardware costs\footnote{In case of LSQ-ResNet18, the baseline quantization accuracy of LSQ is too superior to PACT; thus accuracy gain from NCE cannot overcome this gap.}. 
In particular, NCE outperforms EdMIPS for both ResNet18 and ResNet50, implying the superiority of NCE's search space over EdMIPS's. Furthermore, NCE-ResNet50 achieves the best-to-date 2-bit QDNN performance in terms of accuracy, FLOPs, and PARAMs. This superior performance demonstrates that NCE is a promising solution for compensating for the accuracy degradation of QDNN.

\begin{table}[t]
\begin{center}
\resizebox{\linewidth}{!}{
\begin{tabular}{ccccc}
\Xhline{2\arrayrulewidth}
\multicolumn{5}{c}{ResNet18}    \\ \hline
Method                  & Top-1 Acc(\%)        & Top-5 Acc(\%)        & FLOPs                   & PARAMs                   \\ \Xhline{2\arrayrulewidth}
\textit{Full precision} & \textit{70.56}       & \textit{89.88}       &                         &                         \\ \cline{1-3}
LSQ                     & 67.6                 & 87.6                 & \multicolumn{1}{l}{}    & \multicolumn{1}{l}{}    \\
QIL                     & 65.7                 & -                    & 1.814G                  & 11.69M                  \\
PACT                    & 64.4                 & 85.6                 &                         &                         \\ \cline{1-3}
EdMIPS                  & 65.9                 & 86.5                 & \multicolumn{1}{l}{}    & \multicolumn{1}{l}{}    \\ \cline{1-3}
w/o NCE(Ours)           & 64.08                & 86.47                &                         &                         \\ \hline
\textbf{w/ NCE(Ours)}   & \textbf{66.18}     & \textbf{86.75}       & \textbf{1.897G}         & \textbf{10.94M}         \\ \Xhline{2\arrayrulewidth}
\\ \Xhline{2\arrayrulewidth}
\multicolumn{5}{c}{ResNet50}    \\ \hline
Method                  & Top-1 Acc(\%)        & Top-5 Acc(\%)        & FLOPs                   & PARAMs                   \\ \Xhline{2\arrayrulewidth}
\textit{Full precision} & \textit{76.82}       & \textit{93.33}       &                         &                         \\ \cline{1-3}
LSQ                     & 73.7                 & 91.5                 & \multirow{2}{*}{4.089G} & \multirow{2}{*}{25.56M} \\
PACT                    & 72.2                 & 90.5                 &                         &                         \\ \cline{1-3}
EdMIPS                  & 72.1                 & 90.6                 & \multicolumn{1}{l}{}    & \multicolumn{1}{l}{}    \\ \cline{1-3}
w/o NCE(Ours)           & 72.36                & 90.81                &                         &                         \\ \hline
\textbf{w/ NCE(Ours)}   & \textbf{74.03}       & \textbf{91.63}       & \textbf{3.932G}         & \textbf{17.66M}         \\ \Xhline{2\arrayrulewidth}
\end{tabular}}

\end{center}
\caption{ImageNet: Top-1/Top-5 accuracy and computational complexity for 2-bit (W2A2) QDNN with or without NCE compared with the other state-of-the-art uniform (LSQ, QIL, PACT) and mixed-precision (EdMIPS) quantization methods.}
\label{tab:imagenet}
\end{table}

\subsubsection{Comparison of network structures}
\begin{figure}[t]
\begin{center}
\centerline{\includegraphics[width=0.95\linewidth]{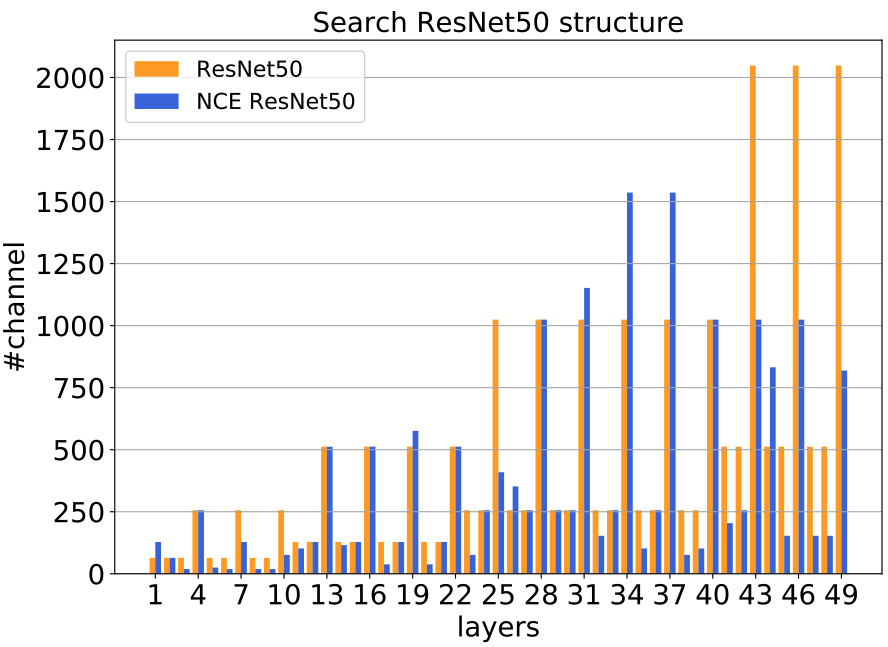}}
\caption{Comparison of structure of ResNet50-ImageNet.}
\label{fig:appendix-comparison-structure-resnset50}
\end{center}
\end{figure}

The neural network searched by NCE has the number of channels expanded or reduced across the layers while maintaining the overall shape. As an example,  Fig.~\ref{fig:appendix-comparison-structure-resnset50} shows the comparison of the number of channels of ResNet50-ImageNet before and after adaptation by NCE. Note that the original ResNet50 structure contains many channels at the last layers, but these numbers of channels are artificially scaled as inversely proportional to the feature-map width. NCE successfully balances the number of channels to improve quantization accuracy while maintaining hardware constraints.

\begin{table}[t]
\begin{center}
\resizebox{0.8\linewidth}{!}{
\begin{tabular}{ccccc}
\Xhline{2\arrayrulewidth}
Threshold $T$ & 0.30    & 0.25    & 0.20    & 0.15    \\ \hline
Accuracy      & 91.60\% & 91.44\% & 91.40\% & 90.77\% \\ \Xhline{2\arrayrulewidth}
\end{tabular}}
\end{center}
\caption{The accuracy of the model NCE adapted from ResNet20-CIFAR10 with different channel preference threshold, $T$.}
\label{tab:appendix-threshold}
\end{table}

\begin{table}[t]
\begin{center}
\resizebox{0.8\linewidth}{!}{
\begin{tabular}{cccc}
\Xhline{2\arrayrulewidth}
\multicolumn{2}{c}{ResNet20-CIFAR10}                                                             & Accuracy & FLOPs                   \\ \hline
\multirow{3}{*}{\begin{tabular}[c]{@{}c@{}}Original structure\\ (w/o NCE)\end{tabular}} & W32A32 & 92.88\%  & \multirow{3}{*}{40.81M} \\
                                                                                        & W3A3   & 92.45\%  &                         \\
                                                                                        & W4A4   & 92.69\%  &                         \\ \hline
\multirow{2}{*}{NCE}                                                                    & W3A3   & 92.66\%  & 39.07M                  \\
                                                                                        & W4A4   & 92.75\%  & 37.07M                  \\ \Xhline{2\arrayrulewidth}
\end{tabular}}
\end{center}
\caption{3-bit and 4-bit Quantization of ResNet20-CIFAR10 with and without NCE.}
\label{tab:appendix-higher-bit-quant}
\vspace{-0.5 cm}
\end{table}

\subsection{Ablation study}

\subsubsection{Channel preference threshold}
The channel preference threshold $T$ in Algorithm~\ref{Algorithm} is the hyper-parameter that determines which layer to request the channel expansion. Given that there are eight search parameters initialized uniformly, $T$ needs to be higher than $1/8=0.125$. Although it plays a critical role in selective channel expansion, the overall performance is not very sensitive to its value. As shown in Table~\ref{tab:appendix-threshold}, there is a plateau in the model accuracy for the selection of $T>0.15$. In practice, setting $T$ between $0.20$ and $0.30$ is sufficient.

\subsubsection{Quantization with higher bit-precision for NCE}
We further experimented with the impact of NCE for higher bit-precision quantization on ResNet20-CIFAR10. As shown in Table~\ref{tab:appendix-higher-bit-quant}, NCE consistently finds neural networks with higher accuracy when they are trained with the same setting as the original network. Interestingly, FLOPs decrease as the bit-precision increases; this indicates that NCE can adequately explore the trade-offs between FLOPs and accuracy.

\section{Conclusion}

In this work, we propose a novel approach that explores the neural network structure to achieve robust inference accuracy while using the simple uniform-precision arithmetic operations. 
Our novel differentiable neural architecture search, called neural channel expansion, employs the search space that can shrink and expand the channels so that the more sensitive layers can be equipped with more channels while the overall resource requirements (e.g., FLOPs and PARAMs) are maintained. Based on in-depth analysis and experiments, we demonstrate that the proposed method can achieve superior performance in ultra-low uniform-precision quantization for CIFAR10 and ImageNet networks.

\begin{acks}
This research was supported by Nota AI. This work was also supported by Institute of Information \& communications Technology Planning \& Evaluation(IITP) grant funded by the Korea government(MSIT) (2020-0-01297, Development of Ultra-Low Power Deep Learning Processor Technology using Advanced Data Reuse for Edge Applications). This work was also supported by the Technology Innovation Program (20013726, Development of Industrial Intelligent Technology for Smart Factory) funded By the Ministry of Trade, Industry \& Energy(MOTIE, Korea)
\end{acks}

\bibliographystyle{named}
\bibliography{ijcai21}

\end{document}